\documentclass{article}
\usepackage{amsmath,graphicx,mlspconf}

\usepackage{multirow}
\usepackage{tabularray}

\usepackage{amssymb}

\usepackage{array}
\usepackage{tablefootnote}
\usepackage{tikz}
\usetikzlibrary{decorations.pathreplacing}
\usetikzlibrary{arrows}

\newcommand{\dd}[2]{\frac{\partial #1} {\partial #2}}

\newcommand{\xa}{x_a}
\newcommand{\xp}{x_p}
\newcommand{\xap}{x_{a,p}}
\newcommand{\cat}{\theta}
\newcommand{\Ra}{r_a}
\newcommand{\Rap}{r_{a,p}}
\newcommand{\Rp}{r_p}

\title{Diffusion models as probabilistic neural operators for recovering unobserved states of dynamical systems}
\name{Katsiaryna Haitsiukevich, Onur Poyraz, Pekka Marttinen, Alexander Ilin \thanks{E-mail: \texttt{\{firstname.lastname\}@aalto.fi} \\
This work was supported by the Research Council of Finland (Flagship programme: Finnish Center for Artificial Intelligence FCAI, and grants 352986, 358246) and EU (H2020 grant 101016775 and NextGenerationEU).\\ We thank Ti John and Nicola Dainese for the help with graphics design.}}
\address{Department of Computer Science, Aalto University, Espoo, Finland}

\copyrightnotice{979-8-3503-7225-0/24/\$31.00 {\copyright}2024 IEEE}

\begin{document}

\maketitle

\begin{abstract}

This paper explores the efficacy of diffusion-based generative models as neural operators for partial differential equations (PDEs). Neural operators are neural networks that learn a mapping from the parameter space to the solution space of PDEs from data, and they can also solve the inverse problem of estimating the parameter from the solution. Diffusion models excel in many domains, but their potential as neural operators has not been thoroughly explored. In this work, we show that diffusion-based generative models exhibit many properties favourable for neural operators, and they can effectively generate the solution of a PDE conditionally on the parameter or recover the unobserved parts of the system. We propose to train a single model adaptable to multiple tasks, by alternating between the tasks during training. In our experiments with multiple realistic dynamical systems, diffusion models outperform other neural operators. Furthermore, we demonstrate how the probabilistic diffusion model can elegantly deal with systems which are only partially identifiable, by producing samples corresponding to the different possible solutions. 

\end{abstract}
\section{Introduction}
\label{sec:intro}

Physical dynamical systems can be expressed with partial differential equations (PDEs) and traditionally solved using methods like the finite difference method. This conventional approach cannot easily incorporate measurements of the process, which motivates the development of neural network methods that can combine the prior knowledge with data-driven learning.
Differential equations can be combined with neural networks by integrating the PDEs directly into the architecture
\cite{greydanus2019hamiltonian, haitsiukevich2021agrid}
or by including the PDE error in the loss to represent prior knowledge 
\cite{raissi2019pinn}.
The first option complies with the underlying physics by design, but is system-specific and non-trivial to extend. In the second approach the architecture does not depend on the equations but the training requires the functional form of the equations and knowledge of the parameters, which may be unavailable.
Alternatively,  it is possible to incorporate domain knowledge through a physics simulator, which can be integrated into the network architecture \cite{yang2022learning} or utilized to augment the training data \cite{lam2023learning}.
The simulator allows to collect a rich data set from the underlying process by randomizing its parameters, which alleviates the need for system identification at inference time, enables a wider class of models, and increases robustness against noise and distortions in the data. If the PDE is known at test time, it can be used to fine-tune the model's prediction \cite{li2021physics}.  

Neural operators are data-driven and learn a mapping from the input parameters or observations to the output. Most of them are deterministic, which suffices for approximating the mapping from the PDE parameters to the solution. However, for the inverse problems of recovering parameters or reconstructing an unobserved state, deterministic predictions may not capture all possible outcomes. To resolve this, we opt for probabilistic generative models, the diffusion models \cite{sohl2015deep, ho2020denoising}, which have excelled in many domains. 
We view the system state evolution, discretized across spacial and temporal dimensions, as an image with state variables stacked as channels. Consequently, the reconstruction of an unobserved state is analogous to a color restoration task. The diffusion models can be pre-trained on complete image data, and conditioning applied only at inference time \cite{lugmayr2022repaint}.
This `unconditional' approach, called RePaint in \cite{lugmayr2022repaint}, is adaptable to many use cases, but with increased inference complexity. Alternatively, the model's input can be augmented with the conditioning information already during training \cite{saharia2022palette, yang2023uni}, yielding a specialized model suitable for a single task. A similar approach has been used for multivariate time series prediction and imputation \cite{alcaraz2023diffusionbased}.

\begin{figure*}[t]
\newcommand{\myheight}{22mm}
\centering
\begin{tblr}{
  colspec = {X[0.19,c,h] X[0.1,r,0.015] X[0.1,r,0.015] | X[0.19,c,h] | X[0.45,c,h] },
  cell{1}{2} = {c=1}{c}, %
  stretch = ,
  rowsep = 1pt,
}
\begin{tabular}{c} Full PDE solution \end{tabular}
&
& \SetCell[r=2]{h} \rotatebox[origin=c]{90}{\small }
& \begin{tabular}{c}Conditional \end{tabular} &  \begin{tabular}{c}Mixed Conditional (M-CEDM)\end{tabular} 
\\
\vspace{-10mm}
\footnotesize \begin{tabular}{c} Desired output \end{tabular} & & & \footnotesize\begin{tabular}{c} \hspace{0mm} Model 1 \hspace{2mm} \dots \hspace{2mm}  Model $5$ \end{tabular} & \footnotesize \begin{tabular}{c} Single model \end{tabular} \\
\\
   \vspace{-30mm}
   \SetCell[r=3]{c}{
   \parbox[t]{24mm}
{

    \hspace{-1mm}
    \begin{tikzpicture}[baseline=-115]
        \draw [->] (-1.4,-5.5) -- (0.1, -5.5) node[below right, pos=0.8] {\footnotesize $t$};
        \draw [->] (-1.4,-5.5) -- (-1.4, -2.3) node[above left, pos=0.9] {\footnotesize $x$};
     \end{tikzpicture} 
     \hspace{-16mm}
     \raisebox{-0.5\height}{\includegraphics[height=27mm,trim={8mm 1mm 1mm 1mm},clip]{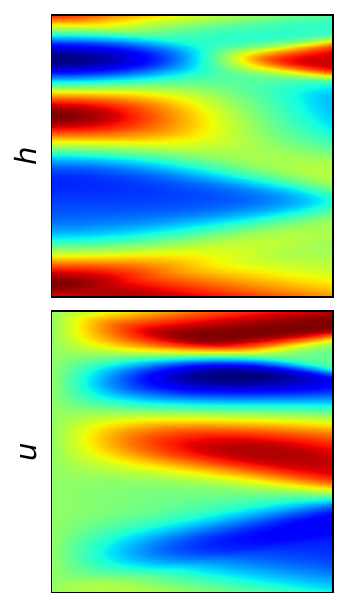}}
     \rotatebox[origin=c]{90}{\hspace{-3mm}\footnotesize {Velocity $u$} \hspace{2mm} \footnotesize {Height $h$}}
   }
   }

  & 
  & \rotatebox[origin=c]{90}{\small Training input}   
  & \includegraphics[height=\myheight,trim={8mm 2mm 2mm 2mm},clip]{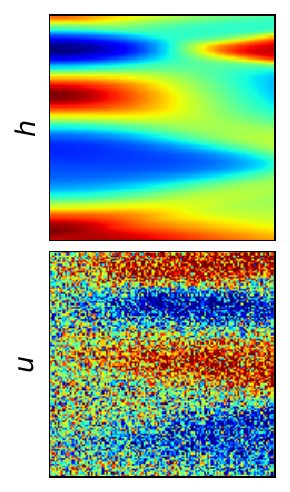} 
  \hspace{2mm}
  \begin{tikzpicture}[baseline=-30]
        \draw [dashed] (1,-1) -- (1,1);
     \end{tikzpicture} 
  \hspace{2mm}
  \includegraphics[height=\myheight,trim={8mm 2mm 2mm 2mm},clip]{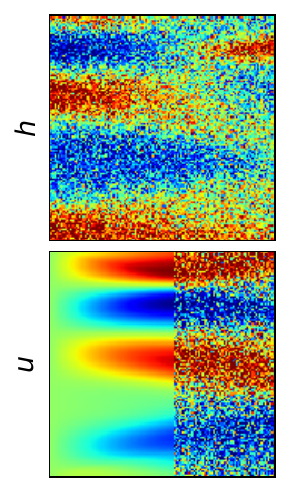}   
  & \includegraphics[height=\myheight,trim={8mm 2mm 2mm 2mm},clip]{img/diff_cases/case1_tr.pdf} 
     \includegraphics[height=\myheight,trim={8mm 2mm 2mm 2mm},clip]{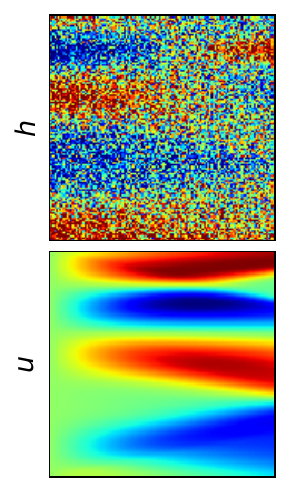}
     \includegraphics[height=\myheight,trim={8mm 2mm 2mm 2mm},clip]{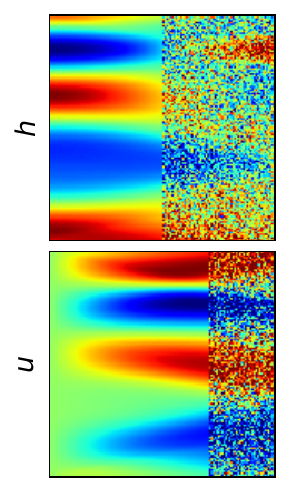}
     \includegraphics[height=\myheight,trim={8mm 2mm 2mm 2mm},clip]{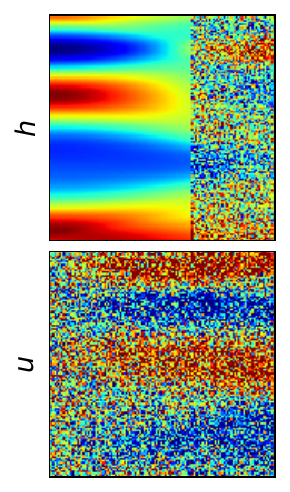}
     \begin{tikzpicture}[baseline=-40]
        \draw [decorate,decoration={brace,amplitude=10pt,mirror,raise=4pt},yshift=0pt] (-1.5,-1.5) -- (-1.5, 0.85) node [black,midway,xshift=0.8cm] {\hspace{5mm} $\thicksim$ \hspace{2mm} };
        \node[text width=1cm] at (-0.3,0.25) {\footnotesize \spaceskip=1pt \parbox{8mm}{each batch}};
        \node[text width=1cm] at (-0.2,-0.5) {\tiny \spaceskip=1pt \parbox{3mm}{i.i.d.}};
        
     \end{tikzpicture}
     \hspace{-5mm}
     \includegraphics[height=\myheight,trim={8mm 2mm 2mm 2mm},clip]{img/diff_cases/case4_tr.pdf}
\vspace{2mm}
\\
& & & \footnotesize\begin{tabular}{c} \hspace{0mm} Task 1 \hspace{2mm} \dots \hspace{3mm}  Task 5 \end{tabular} & \footnotesize\begin{tabular}{c} \hspace{-2mm} Task 1 \hspace{4.5mm} Task 2 \hspace{4.5mm} Task 3 \hspace{4.5mm} Task 4 \hspace{4.5mm} Task 5 \end{tabular} \\
\vspace{-35mm}
\\

  &
  &
  \rotatebox[origin=c]{90}{\small  Inference input}
  &  \includegraphics[height=\myheight,trim={2mm 2mm 2mm 2mm},clip]{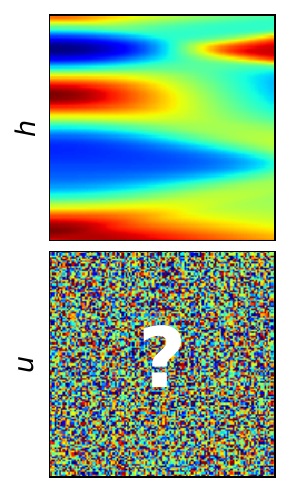} 
  \hspace{2mm}
  \begin{tikzpicture}[baseline=-30]
        \draw [dashed] (1,-1) -- (1,1);
     \end{tikzpicture} 
  \hspace{2mm}
  \includegraphics[height=\myheight,trim={8mm 2mm 2mm 2mm},clip]{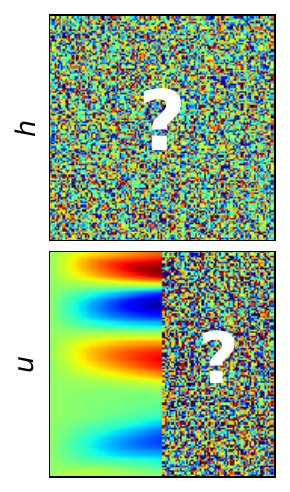} 
  & \includegraphics[height=\myheight,trim={8mm 2mm 2mm 2mm},clip]{img/diff_cases/case1_inf.pdf}
     \hspace{1mm}
     \includegraphics[height=\myheight,trim={8mm 2mm 2mm 2mm},clip]{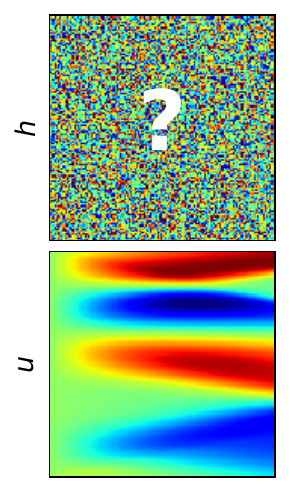}
     \hspace{1mm}
     \includegraphics[height=\myheight,trim={8mm 2mm 2mm 2mm},clip]{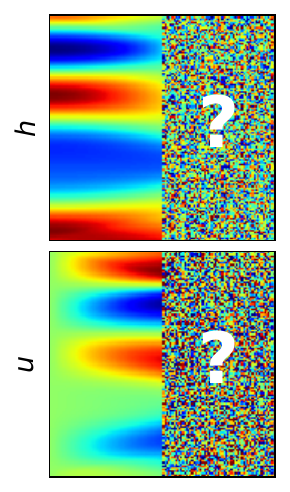}
     \hspace{1mm}
     \includegraphics[height=\myheight,trim={8mm 2mm 2mm 2mm},clip]{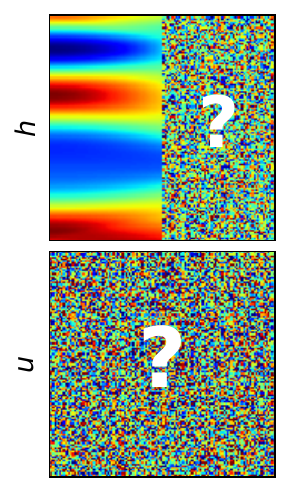}
     \hspace{1mm}
     \includegraphics[height=\myheight,trim={8mm 2mm 2mm 2mm},clip]{img/diff_cases/case5_inf.pdf}

\end{tblr}
\caption{Conditional and mixed conditional training of diffusion models demonstrated with the Shallow-Water Equation (SWE-orig). The system has two variables (channels) represented by the colored rectangles with time ($t$) and spatial coordinate ($x$) on the x- and y-axes. The `desired output' is the full state which we train the model to reconstruct from partial information. The clean parts of the input represent the conditioning information (at training and inference time) and the noisy parts are reconstructed by denoising.
Each \textbf{conditional} model is trained with conditioning information for a single task. %
The \textbf{mixed conditional} training yields a single model for all defined tasks by sampling one task for each mini-batch during training.
}
\label{f:method_overview}
\end{figure*}

Our novel contributions are: \textbf{1)} We define and compare multiple diffusion models for unobserved state reconstruction and prediction. Unlike existing works in this domain \cite{li2021fourier,li2023transformer}, 
we train our model with a `mixed conditional' objective, allowing it to solve multiple tasks with a single training (see Fig.~\ref{f:method_overview}).
\textbf{2)} We empirically demonstrate the strong performance of diffusion models in general, and the model with mixed conditional training in particular, against other neural operator baselines. \textbf{3)} We study systems which are not fully identifiable, and demonstrate that diffusion models can represent the variability of possible solutions. 

\section{Related work}
\label{sec:related_work}

{\bf Neural operators} \cite{kovachki2023neural} have strong performance in modelling of physical systems. State-of-the-art results on rectangular domains are produced by FNO \cite{li2021fourier}. Recently proposed neural operators use the transformer architecture, e.g., OFormer \cite{li2023transformer}. 
Existing methods solve either the forward (parameters to solution) or the inverse (solution to parameters) problem. Unlike the neural operator approaches, a diffusion model can solve both of these problems at once by conditioning with known information. Concurrent to our work, \cite{long2024invertible} combines forward and inverse operators but in a non-diffusion model. 

{\bf Diffusion models for physical systems} have been successfully applied for predicting the dynamics of the system in the future, either by autoregressive unrolls \cite{kohl2023turbulent} or by taking the desired future time as a model input \cite{yang2023denoising}.
These models focus on prediction only while we consider training jointly for prediction and reconstruction. A work \cite{apte2023diffusion} utilized the diffusion model for data generation for several physical systems. It focused on unconditional generation that adheres to physical constraints of the system while we recover the unobserved states conditioned on partial information.

{\bf Incorporation of PDE prior information into diffusion models}.
A physics-informed diffusion model \cite{bastek2024physics} includes  a scaled PDE residual into the training objective, encouraging solutions that comply with the PDE. Alternatively, the gradient of the PDE residual guide the sampling \cite{jacobsen2023cocogen} or be directly fed into the model as an extra input channel during training \cite{shu2023physics}. 
All of the approaches can be applied together with the methods in this paper.

\section{Method}

\begin{figure}[t]
\newcommand{\myheight}{13mm}
\centering
\begin{tblr}{
  colspec = {X[0.605,h]},
  cell{1}{2} = {c=1}{c}, %
  stretch = 0,
  rowsep = 0pt,
}
\hspace{-3mm} \begin{tabular}{c} \small States to reconstruct\end{tabular} \hspace{2mm} \begin{tabular}{c} \small Conditioning\end{tabular} \hspace{2mm} \begin{tabular}{c} \small Binary masks\end{tabular} \\
\vspace{-15mm}
\\
\includegraphics[height=\myheight,trim={2mm 2mm 2mm 2mm},clip]{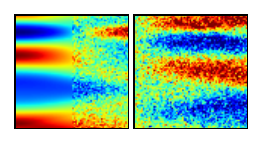}
\includegraphics[height=\myheight,trim={2mm 2mm 2mm 2mm},clip]{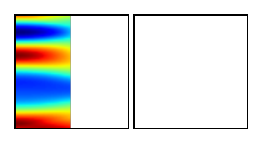}
\includegraphics[height=\myheight,trim={2mm 2mm 2mm 2mm},clip]{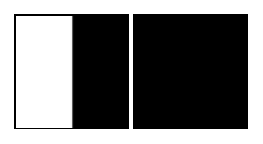}

\end{tblr}
\caption{Model input for mixed conditional training.
}
\label{f:model_input}
\end{figure}

\subsection{Background on diffusion models}

Denoising diffusion probabilistic models (DDPMs) \cite{ho2020denoising} are a class of generative models which learn underlying data distribution by progressively removing noise from the distorted input. In particular, the training procedure requires defining a schedule for the magnitude of noise added to model input at every training iteration. The highest noise magnitude in the schedule makes the model input close to pure Gaussian noise. A neural network, predicting a denoised version of the input, is trained by optimizing a variational lower bound or a simplified objective \cite{ho2020denoising} between the added noise and the network prediction, or some other refined combinations (see, e.g., \cite{nichol2021improved, karras2022elucidating}).
Denoising diffusion models use the U-net architecture \cite{ronneberger2015u} with conditioning on the step $t$ of the noise schedule \cite{ho2020denoising}. The original architecture has been made more expressive by improvements in residual blocks \cite{song2020score, dhariwal2021diffusion} while inefficiencies in training dynamics have been remedied by input and output preconditioning \cite{karras2022elucidating}.

One downside of diffusion models is the slow inference, which, starting from random noise, iteratively denoises the input into the final prediction. The original DDPM \cite{ho2020denoising} sampling uses the same number of steps as the training noise schedule (typically around 1000 steps). However, the number of inference steps can be significantly decreased by defining a separate inference schedule \cite{nichol2021improved,song2020denoising,karras2022elucidating}. The approaches \cite{nichol2021improved,song2020denoising} are uniformly skipping some steps in the training noise schedule. However, a better strategy for allocating the inference resources is skipping more steps in the beginning of the sampling process and preserving more steps closer to the end where the noise magnitude is smaller. This idea combined with a second order Heun method has been proposed by \cite{karras2022elucidating} and together with input and output preconditioning is referred to as EDM.

\subsection{Mixed conditional training}
We aim at training a model capable of solving several tasks, such as the unobserved state reconstruction, the system dynamics prediction from the parameter value, the underlying parameter recovery from the dynamics, and the prediction of future system states. The parameter values can be concatenated with the system states as an additional input channel making parameter recovery and dynamics prediction analogous to the state reconstruction task.
Similar problem formulation arises in image-to-image translation considered in Palette \cite{saharia2022palette} and in time series imputation and forecasting addressed in SSSD\textsuperscript{S4} \cite{alcaraz2023diffusionbased}. 
We propose to model the dynamics by a conditional diffusion model trained on a mixture of the tasks of interest as shown in Fig.~\ref{f:method_overview}.

To simultaneously train for several tasks, we generate a mask corresponding to the unknown part of the system (e.g. upper or lower channel in Fig.~\ref{f:method_overview}) for each training batch. For the observed channel(s), we sample the number of time steps observed (in the direction of the x-axis in Fig.~\ref{f:method_overview}). For example, Task 3 in Fig.~\ref{f:method_overview} corresponds to observing both channels partially for the first half of the time span, while in Tasks 4 and 5 only one channel is partially observed.

Since a part of the full state is known, the diffusion process is applied only to the unobserved part. As shown in Fig.~\ref{f:model_input}, the input for one denoising step consists of system states, such that noise is added to the unobserved part, and the observed part is kept noise free and given as conditioning information. The input is the same both at training and inference times. The conditioning information can be optionally accompanied by the corresponding binary masks as in SSSD\textsuperscript{S4} which we also found beneficial. To take into account that the denoising should be applied only to the unknown part, the observed part is masked out from the loss.

\section{Experiments}

\begin{figure}[t]
\newcommand{\myheight}{36mm}
\centering
\begin{tblr}{
  colspec = {X[0.05,r,0.015]X[0.17,c,h]X[0.05,r,0.015]X[0.30,h]},
  cell{1}{2} = {c=1}{c}, %
  stretch = 0,
  rowsep = 0pt,
}
  \rotatebox[origin=c]{90}{\small (a) Darcy flow}  
  & \hspace{-4mm} \includegraphics[height=\myheight,trim={1mm 1mm 1mm 1mm},clip]{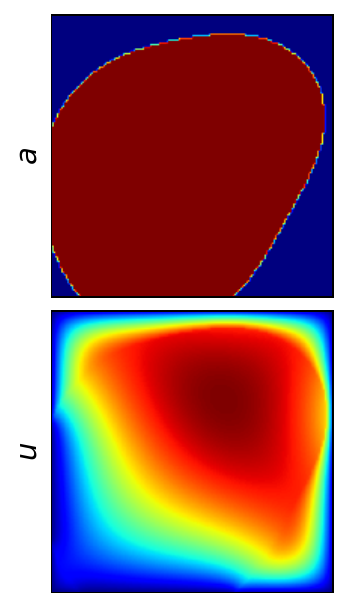}
  & \rotatebox[origin=c]{90}{\small (b) Reactor} 
  & \includegraphics[height=\myheight,trim={1mm 1mm 1mm 1mm},clip]{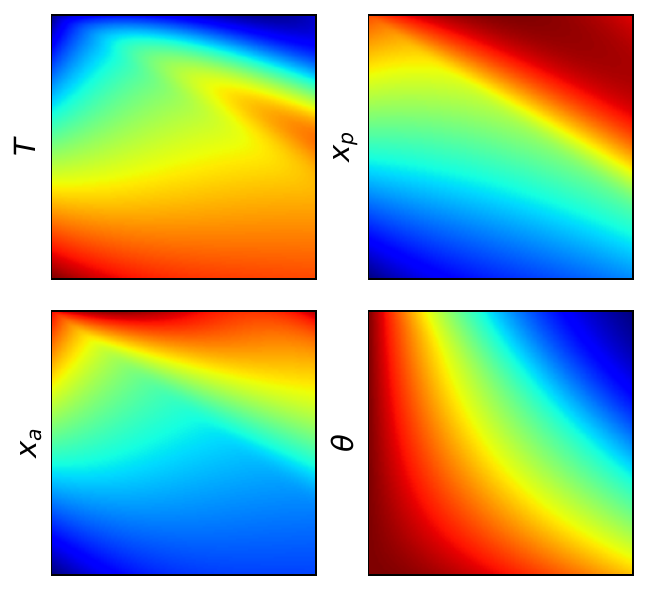}

\end{tblr}
\vskip -2mm
\caption{Example simulations from two of the studied systems.
}
\label{f:systems}
\end{figure}

\begin{table*}[t]
\caption{
Model comparison on state reconstruction for the forward (Task 1, columns 2 -- 4) and inverse (Task 2, columns 5 -- 7) problems in case of identifiable systems. For each problem we present the mean absolute error (MAE) and the PDE residual error. All MAE values should be multiplied by $10^{-3}$.
}
\label{tab:res_state1}
\begin{center}
\begin{tabular}{@{}c|cc|cc|cc||cc|cc|cc}
\multirow{3}{*}{Model}   & \multicolumn{2}{c|}{\multirow{2}{*}{Darcy, $a \rightarrow u$}} & \multicolumn{2}{c|}{\multirow{2}{*}{SWE, $h \rightarrow u$}} & \multicolumn{2}{c||}{Reactor,} & \multicolumn{2}{c|}{\multirow{2}{*}{Darcy, $u \rightarrow a$}} & \multicolumn{2}{c|}{\multirow{2}{*}{SWE, $u \rightarrow h$}} & \multicolumn{2}{c}{Reactor,}
\\
& & & & & \multicolumn{2}{c||}{$\xp, T \rightarrow \theta, \xa$} & & & & & \multicolumn{2}{c}{$\theta, \xa \rightarrow \xp, T$}
\\
\cline{2-13}
                         & \textit{MAE} & \textit{PDE} & \textit{MAE} & \textit{PDE} & \textit{MAE} & \textit{PDE} & \textit{MAE}  & \textit{PDE} & \textit{MAE} & \textit{PDE} & \textit{MAE} & \textit{PDE}
\\
\hline
FNO~\cite{li2021fourier}                     & 7.55 & 38.08 & 0.44 & 0.9963 & 0.07 & 0.0146 & 3.38 & 3.236 & 2.21 & 1.1370 & 5.64 & 0.1132 \\
OFormer~\cite{li2023transformer}             & 6.82 & 21.00 & 0.31 & 0.9966 & 0.13 & 0.0083 & 4.06 & 6.683 & 0.81 & 0.9929 & 7.76 & 0.0282 \\
RePaint~\cite{lugmayr2022repaint}            & 9.80 & 6.909 & 0.69 & 1.0462 & 0.73 & 0.1521 & 191.8 & 84.11 & 9.68 & 1.0429 & 188.3 & 0.2017 \\
CDDPM~\cite{ho2020denoising}                 & 6.23 & 1.478 & 0.25 & 0.9925 & 0.09 & 0.0039 & 0.25 & \textbf{1.180} & 0.84 & \textbf{0.9906} & 4.62 & 0.0039 \\
CEDM~\cite{karras2022elucidating}            & 5.52 & 1.344 & \textbf{0.10} & 0.9923 & \textbf{0.05} & \textbf{0.0015} & 0.18 & 1.192 & 0.44 & 0.9924 & 3.42 & \textbf{0.0017} \\
\hline
M-CEDM              & \textbf{5.46} & \textbf{1.272} & \textbf{0.10} & \textbf{0.9915} & 0.07 & 0.0020 & \textbf{0.17} & \textbf{1.180} & \textbf{0.43} & \textbf{0.9906} & \textbf{2.69} & \textbf{0.0017} \\
\end{tabular}
\end{center}
\end{table*}

\subsection{Dynamical systems considered}
We demonstrate the performance of the proposed diffusion models on four systems. The first three: Darcy flow, shallow water equation with periodic initial conditions, and the fixed-bed tubular reactor are fully identifiable given the conditioning information. The fourth system is the shallow water equation with an initial perturbation of the water level, and it is partially non-identifiable, meaning that multiple solutions are compatible with the given conditioning information. Below we specify these systems and their data generation.

\textbf{Darcy Flow equation} is defined over a unit square with a 2D steady-state solution governed by
\begin{align}
     \nabla \left(a(x) \nabla u(x)\right) &= f(x) \text{, } x \in (0, 1)^2, \\
     u(x) &= 0 \text{, } x \in \partial(0, 1)^2,
    \label{eq:darcy}
\end{align}
with viscosity $a(x)$ being piece-wise constant and the force term $f(x) =1$. We use a subset of 1000 simulations from PDEBench \cite{takamoto2022pdebench}, of which one instance is shown in Fig.~\ref{f:systems}(a).

\textbf{Shallow water equation (SWE-orig)} is derived from the compressible Navier-Stokes equation and describes free-surface fluid flow. The system is characterized by height $h$ and velocity $u$ of the fluid evolving as:
\begin{align}
    \dd{h}{t} &= -\dd{hu}{x} \text{, } \quad \dd{hu}{t} = -\dd{(u^2h + 0.5gh^2)}{x}
    \label{eq:swe}
\end{align}
where $g$ is the gravitational constant. Here, we set $g=1$ and randomize the initial conditions for $h$. For $t \in [0, 0.128]$, $x \in [-0.5, 0.5]$, and  $u_0=0$, $h_0$ is calculated as:
\begin{align}
    \tilde{h}(0, x) &= \sum_{k=-N}^{N} \lambda_k \cos(2 \pi k x) + \gamma_k \sin(2\pi k x), \\
    h(0, x) &= 1 + \frac{\tilde{h}(0, x) - \min(\tilde{h}(0, x))}{\max(\tilde{h}(0, x)) - \min(\tilde{h}(0, x))} \text{,}
    \label{eq:swe_ic_period}
\end{align}
where $N=3$ and $\lambda_k, \gamma_k \sim \mathcal{N}(0, 1)$. An example of this system is used in the illustrations in Fig.~\ref{f:method_overview}.

\textbf{Shallow water equation (SWE-init)} is generated for $x \in [-2.5, 2.5]$ and $t \in [0, 1.28]$ with initial conditions:
\begin{align}
    h(0, x) &= h_\text{in} + \epsilon \exp\left(-\frac{(x - x_0)^2}{2 \sigma^2}\right) \text{, } \quad
    u(0, x) = u_0 \text{,}
    \label{eq:swe_ic}
\end{align}
where $h_\text{in} \sim \mathcal{U}\left(1.2, 5.2\right)$, $\epsilon \sim \mathcal{U}\left(0.05, 1\right)$, $x_0 \sim \mathcal{U}\left(-1, 1\right)$, $\sigma \sim \mathcal{U}\left(0.2, 2\right)$,  $(hu)_0 \sim \mathcal{U}\left(-2.2, 2.2\right)$. This equation models an abrupt change in the water level, e.g. caused by a dam break. Depending on the initial water velocity $u_0$, several possible outcomes may correspond to the same observation of the height $h$ (Fig.~\ref{f:swe_predict}).

For both SWE datasets the data are simulated using PyClaw \cite{ketcheson2012pyclaw} Python package implementing a finite volume method for this equation.

\textbf{Fixed-bed tubular reactor} is a system with hydrogenation of aromatics with concentration $\xa$ on the catalyst surface with activity $\cat$. The reactor's input contains a poisoning agent with concentration $\xp$ which deteriorates activity $\cat$ \cite{price1977catalyst}. The system dynamics in space $z$ and time $t$ is modelled as
\begin{align} 
\dd{\xap}{t} &= -U \dd{\xap}{z} - \alpha(T) \Rap, \\
\dd{T}{t} &= - \beta(\xa, T) U \dd{T}{z} + \gamma \Ra, \quad
\frac{d \theta}{dt} = -r_d,
\label{eq:reactor}
\end{align}
where $U$ is the fluid velocity, $\alpha(T)$ and $\beta(\xa, T)$ are functions of temperature $T$ and the concentration $\xa$, $\gamma$ is a constant and the functions $\Ra$, $\Rp$ and $r_d$ are reaction, poisoning adsorption, and catalyst deactivation rates (see \cite{haitsiukevich2021agrid} for more details). A simulation of the system is shown in Fig.~\ref{f:systems}(b).

\subsection{Baselines}
We compared the proposed approach against the (non-diffusion) supervised neural operators FNO \cite{li2021fourier} and OFormer \cite{li2023transformer}, which have been trained using the implementations provided by the authors.
As diffusion model baselines, we compared the mixed conditional training (M-CEDM) with the `unconditional' RePaint approach and the conditional diffusion models based on the DDPM (CDDPM) and EDM (CEDM). 
RePaint and CDDPM are trained on the simplified objective \cite{ho2020denoising} by predicting added noise, while CEDM and M-CEDM use the EDM objective \cite{karras2022elucidating} and predict a mixture of noise and denoised image. All diffusion models rely on the 2nd order Heun sampler \cite{karras2022elucidating}. In all experiments we take 100 samples from each diffusion model and calculate the mean absolute error (MAE) and the PDE residual error by averaging across those samples. The number of parameters in the different models is approximately equal. The code and the datasets will be released upon acceptance.

\subsection{Reconstruction of the unobserved variables}
In this experiment (results in Table~\ref{tab:res_state1}), we consider two tasks: forward and inverse. In the forward task we predict the solution $u$ conditioned on the parameter $a$ (Darcy), the velocity $u$ from the height $h$ (SWE-orig) and both the concentration $\xa$ and the catalyst $\cat$ from the temperature $T$ and poisoning concentration $\xp$ (Reactor). For the inverse task the inputs and the targets are flipped (Tasks 1, 2 in Fig.~\ref{f:method_overview}). For these tasks the supervised neural operators, FNO and OFormer, provide strong results. However, the PDE residuals are consistently smaller for the conditional and mixed conditional diffusion models (CDDPM, CEDM and M-CEDM), which means that their solutions are more compatible with the prior knowledge of the system dynamics. Since CEDM results were better than CDDPM, we selected it as the basis for training the mixed conditional approach. Comparing RePaint trained `unconditionally' to generate both states jointly against the specialized conditional models, we conclude that the conditioning during training is beneficial. However RePaint uses a singe model for both forward and inverse problems, while the conditional models (including FNO and OFormer) use a dedicated model for each task. Similar to RePaint, M-CEDM uses a single model for both tasks but with performance similar to or better than the other conditionally trained models.

\subsection{Future state prediction}
Next, we compared the methods on Task 3 (future prediction) and Tasks 4-5 (reconstruction and prediction) in Fig.~\ref{f:method_overview}, results shown in Table~\ref{tab:res_time_pred}. For RePaint and M-CEDM, we used the same trained model as in Table~\ref{tab:res_state1}. We selected M-CEDM as a representative of the conditional diffusion models as it was the best in the previous experiment and did not require retraining. For FNO and OFormer we had two models, one to reconstruct the state, the other for prediction. Both FNO and OFormer can handle inputs with different amounts of observed time steps at test time; however, the performance drops significantly when the models trained on the full state are applied to partial observations (marked by * in Table \ref{tab:res_time_pred}). By retraining FNO and OFormer for the correct input size, the performance improved but stayed below the diffusion model.

\begin{table}[t]
\caption{
Results for SWE-orig dataset in reconstruction and prediction (Tasks 3-5). MAE values should be multiplied by $10^{-3}$. Models with * are the same as in Table \ref{tab:res_state1} and trained for full state reconstruction.
}
\label{tab:res_time_pred}
\begin{center}
\begin{tabular}{@{}c|p{0.5cm}c|p{0.5cm}c|p{0.5cm}c}
\multirow{2}{*}{Model}                         & \multicolumn{2}{c|}{Task 3} & \multicolumn{2}{c|}{Task 4} & \multicolumn{2}{c}{Task 5}
\\
                                               & \textit{MAE} & \textit{PDE} & \textit{MAE} & \textit{PDE} & \textit{MAE} & \textit{PDE} 
\\
\hline
FNO*~\cite{li2021fourier}                      & 0.70 & \textbf{0.98} & 51.2 & 47.5 & 90.7 & 150 \\
OFormer*~\cite{li2023transformer}              & 0.40 & 1.00 & 30.3 & 105 & 92.8 & 47.5 \\
FNO~\cite{li2021fourier}                       & 0.70 & \textbf{0.98} & 0.56 & \textbf{0.98} & 1.34 & 1.04 \\
OFormer~\cite{li2023transformer}               & 0.40 & 1.00 & 0.35 & 1.00 & 1.04 & 1.01 \\
RePaint~\cite{lugmayr2022repaint}              & 0.65 & 1.08 & 0.79 & 1.04 & 54.2 & 4.72 \\
\hline
M-CEDM                                         & \textbf{0.08} & 0.99 & \textbf{0.14} & 0.99 & \textbf{0.48} & \textbf{0.99}  \\
\end{tabular}
\end{center}
\end{table}

\subsection{Recovering several possible outcomes}

\begin{table}[t]
\caption{
Results of state reconstruction (Task 1) in the non-identifiable case (SWE-init). All MAE values should be multiplied by $10^{-3}$. We see that samples from the diffusion model include the correct prediction (`closest'), and additional prior information (extra observed points or PDE error) can be useful to identify the most plausible samples.
}
\label{tab:res_unidentifiable}
\begin{center}
\begin{tabular}{@{}l|cc}
\hspace{2cm} Model                                          & \textit{MAE} & \textit{PDE}  
\\
\hline
Kalman Filter~\cite{kalman1960new}                          & 52.2 & 0.2858 \\
FNO~\cite{li2021fourier}                                    & 2.43 & 0.1300 \\
OFormer~\cite{li2023transformer}                            & 2.74 & 0.0400 \\
\hline
Diffusion model, mean prediction                            & 55.1 & 0.0193 \\
 \hspace{0.2cm} - prediction selected by PDE error          & 13.7 & \textbf{0.0159} \\
 \hspace{0.2cm} - prediction selected by 2 corner points    & 1.92 &  0.0161 \\
 \hspace{0.2cm} - prediction closest to the target          & \textbf{1.53} & 0.0162 \\
\end{tabular}
\end{center}
\end{table}

Here we compare the models for velocity reconstruction (Task 1) in the SWE-init system, and the results are summarized in Table~\ref{tab:res_unidentifiable}. The system is partially non-identifiability, which means that without additional information multiple reconstructions can be compatible with the conditioning information. The purpose of this experiment is to demonstrate the benefits of the probabilistic diffusion models in this scenario, namely their ability to generate multiple different samples reflecting the variability of possible solutions, and including the correct target value. Here, we include also the Kalman Filter \cite{kalman1960new}, obtained by linearizing the PDEs~\cite{rafiee2009kalman}, as another probabilistic baseline.

\begin{figure}[t]
\newcommand{\myheight}{62mm}
\centering
\includegraphics[height=\myheight,trim={2mm 3mm 2mm 2mm},clip]{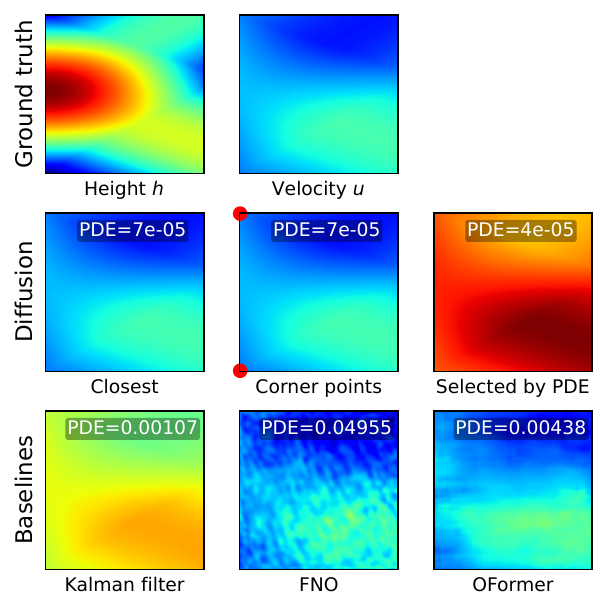}

\vskip -3mm
\caption{State reconstruction results for non-identifiable system SWE-init for a test sample with multiple possible outcomes. }
\label{f:swe_predict}
\end{figure}

As shown in Table~\ref{tab:res_unidentifiable}, the diffusion model mean prediction, obtained by averaging over the samples, has a higher MAE than the other methods, as expected. To illustrate this further, Fig.~\ref{f:swe_predict} presents results for one test example with a large mean MAE. We see that the diffusion model produced samples which all have the same overall shape, but which differ in the initial velocity, which reflects true non-determinacy of the system. Further, the correct result is among the samples (`closest'), and just by including two additional points in the conditioning set it can already be identified. The deterministic FNO and OFormer models have a relatively small MAE but they also introduce artifacts, making the prediction not compatible with the physical dynamics, as seen also in the large PDE-errors. The Kalman filter yields a plausible prediction but is less precise due to the linearization. 

\begin{figure}[t]
\newcommand{\myheight}{27mm}
\centering
\vskip -2mm
\includegraphics[height=\myheight,trim={2mm 3mm 2mm 2mm},clip]{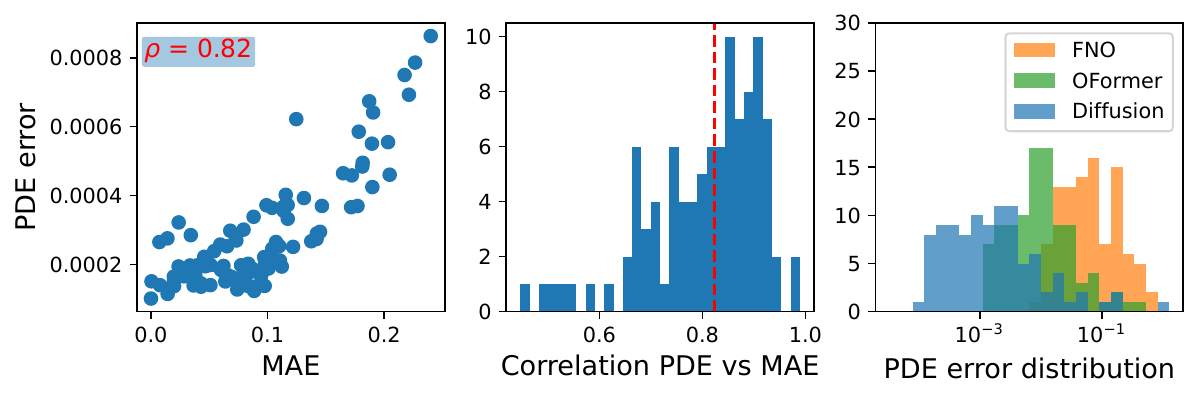}

\vskip -2mm
\caption{Detailed results for the non-identifiable system SWE-init. \textbf{Left}: PDE error vs. MAE for 100 samples for a single test case. \textbf{Middle}: Histogram of correlations between MAE and PDE error across the test set (red line shows the case on the left). \textbf{Right}: PDE errors of different models in the test set. }
\label{f:swe_mae_pde_correlation}
\end{figure}

Finally, Fig.~\ref{f:swe_mae_pde_correlation} highlights the potential of the PDE-error as a prior to identify the most plausible samples. For all test examples the PDE-error is clearly correlated with the MAE (left and middle panels in Fig.~\ref{f:swe_mae_pde_correlation}). However, even among the samples with the smallest PDE-errors, there can still be variability, as demonstrated on the middle row in Fig.~\ref{f:swe_predict}. The rightmost panel in Fig.~\ref{f:swe_mae_pde_correlation} confirms the earlier finding that the diffusion model produces samples with a smaller PDE-error, meaning that they better comply with the underlying physics.

\section{Discussion and Future work}

We proposed a diffusion model that can be used as a neural operator to solve several tasks without retraining. Additionally, when the problem is not fully defined, the probabilistic model, unlike the deterministic alternatives, can produce several possible results compatible with the conditioning information, and we studied using  prior knowledge, either additional measured points or the PDE error, as a way of selecting the most plausible samples. 
As a limitation, the inference with the diffusion model requires sampling, which, despite recent improvements, is still slower than with the deterministic baselines. As future work, the model architecture could be tailored for the specifics of the dynamical systems.

\bibliographystyle{IEEEbib}
\small
\bibliography{ref_short}

\end{document}